\documentclass[10pt,twocolumn,letterpaper]{article}

\usepackage{iccv}
\usepackage{times}
\usepackage{epsfig}
\usepackage{graphicx}
\usepackage{amsmath}
\usepackage{amssymb}
\usepackage[accsupp]{axessibility}  

\usepackage[breaklinks=true,bookmarks=false]{hyperref}

\iccvfinalcopy 

\def\iccvPaperID{10338} 

\ificcvfinal\fi

\graphicspath{{figures/}}

\def\exp{\text{exp\,}}
\def\eg{\emph{e.g.}\,}
\def\ie{\emph{i.e.}\,}

\usepackage{multirow}
\usepackage{booktabs}
\usepackage{arydshln}
\usepackage{pifont}
\newcommand{\ra}[1]{\renewcommand{\arraystretch}{#1}}

\begin{document}

\title{Vision Grid Transformer for Document Layout Analysis}

\author{Cheng Da\thanks{Equal contribution. $\dagger$ Corresponding author.} , Chuwei Luo$^{*}$, Qi Zheng, and Cong Yao$^{\dagger}$  \\
DAMO Academy, Alibaba Group, Beijing, China\\
{\tt\small {dc.dacheng08,luochuwei,zhengqisjtu,yaocong2010}@gmail.com}
}

\maketitle
\ificcvfinal\thispagestyle{empty}\fi

\begin{abstract}
Document pre-trained models and grid-based models have proven to be very effective on various tasks in Document AI. However, for the document layout analysis (DLA) task, existing document pre-trained models, even those pre-trained in a multi-modal fashion, usually rely on either textual features or visual features. Grid-based models for DLA are multi-modality but largely neglect the effect of pre-training. To fully leverage multi-modal information and exploit pre-training techniques to learn better representation for DLA, in this paper, we present VGT, a two-stream Vision Grid Transformer, in which Grid Transformer (GiT) is proposed and pre-trained for 2D token-level and segment-level semantic understanding. Furthermore, a new dataset named D$^4$LA, which is so far the most diverse and detailed manually-annotated benchmark for document layout analysis, is curated and released. Experiment results have illustrated that the proposed VGT model achieves new state-of-the-art results on DLA tasks, e.g. PubLayNet ($95.7\%$$\rightarrow$$96.2\%$), DocBank ($79.6\%$$\rightarrow$$84.1\%$), and D$^4$LA ($67.7\%$$\rightarrow$$68.8\%$). The code and models as well as the D$^4$LA dataset will be made publicly available \footnote{https://github.com/AlibabaResearch/AdvancedLiterateMachinery}.
\end{abstract}

\section{Introduction}

Documents are important carriers of human knowledge.
With the advancement of digitization, the techniques for automatically reading~\cite{Shi2015AnET,Zhou2017EASTAE,Shi2019ASTERAA,mgp,levocr}, parsing~\cite{Zhu2015SceneTD,Long2018SceneTD}, and understanding documents~\cite{Luo_2023_CVPR,Yang2023ModelingEA,huang2022layoutlmv3,li2022dit} have become a crucial part of the success of digital transformation~\cite{cui2021document}.
Document Layout Analysis (DLA)~\cite{binmakhashen2019document}, which transforms documents into structured representations, is an essential stage for downstream tasks, such as document retrieval, table extraction, and document information extraction.
Technically, the goal of DLA is to detect and identify homogeneous document regions based on visual cues and textual content within the document.
However, performing DLA in real-world scenarios is faced with numerous difficulties: 
variety of document types, complex layouts, low-quality images, semantic understanding, etc.
In this sense, DLA is a very challenging task in practical applications.

\begin{table}[t] \centering
\setlength{\tabcolsep}{4pt}
\ra{1.2}
\caption{Comparisons of the use of different modalities and pre-training techniques with existing SOTA methods in DLA task.}
\label{tab:intro_model_case}
\begin{tabular}{cccc}
   \toprule[1pt]
   \textbf{Models} & \textbf{Vision} & \textbf{Text} & \textbf{Pre-trained} \\
   \midrule
   CNN-based \cite{schreiber2017deepdesrt,CascadeTabNet} & \ding{51} & \textcolor[rgb]{1,0,0}{\ding{55}}  &\textcolor[rgb]{1,0,0}{\ding{55}} \\ 
   \hline
   ViT-based \cite{li2022dit} & \ding{51} & \textcolor[rgb]{1,0,0}{\ding{55}} & \ding{51} \\ 
   \hline
   Multi-modal PTM \cite{gu2021unidoc, huang2022layoutlmv3} & \ding{51} & \textcolor[rgb]{1,0,0}{\ding{55}}          & \ding{51} \\
   \hline
   Grid-based \cite{yang2017learning,zhang2021vsr} & \ding{51} & \ding{51} & \textcolor[rgb]{1,0,0}{\ding{55}}  \\ 
   \hline
   \textbf{VGT (Ours)}      & \ding{51} & \ding{51} & \ding{51} \\ 
   \bottomrule[1pt]
\end{tabular}
\end{table}

\begin{figure*}[!htp]\centering
 \includegraphics[width=0.95\textwidth]{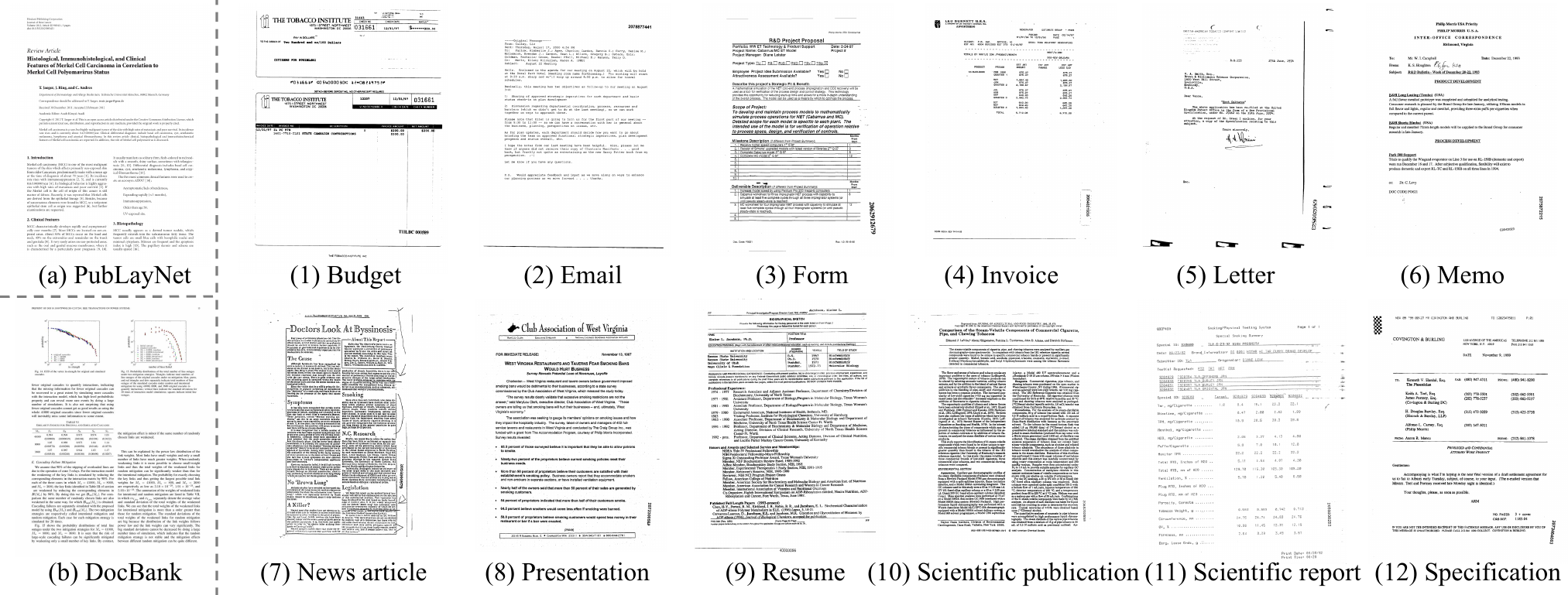}
\caption{Document examples in the public dataset PubLayNet (a) and DocBank (b) and document examples in real-world applications.}
 \label{fig:intro_case}
\end{figure*}

\begin{figure}[!htp]\centering
 \includegraphics[width=0.48\textwidth]{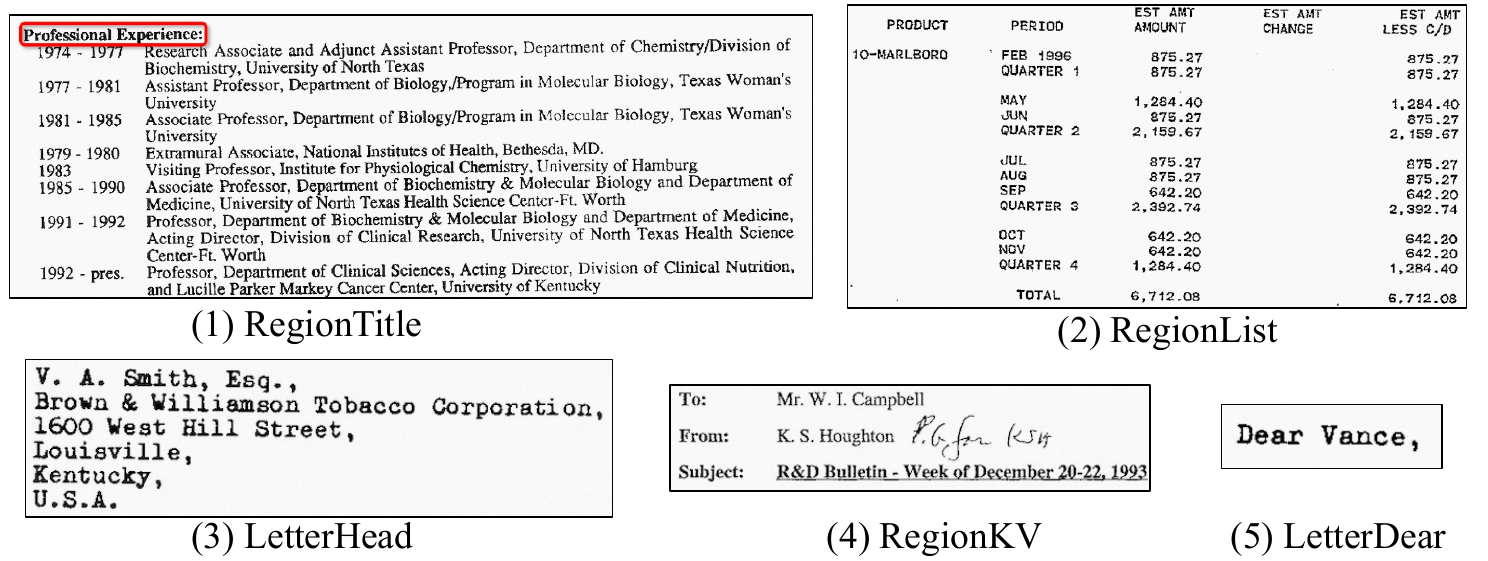}
\caption{Some special layout categories of D$^4$LA dataset.}
 \label{fig:vis_cats}
\end{figure} 

Basically, DLA can be regarded as an object detection or semantic segmentation task for document images in computer vision.
Early works~\cite{schreiber2017deepdesrt,CascadeTabNet} directly use visual features encoded by convolutional neural networks (CNN)~\cite{he2016deep} for layout units detection~\cite{ren2015faster,liu2016ssd, redmon2018yolov3, he2017mask}, and have been proven to be effective.
Recent years have witnessed the success of document pre-training.
Document Image Transformer (DiT)~\cite{li2022dit} uses images for pre-training, obtaining good performance on DLA.
Due to the multi-modal nature of documents, previous methods~\cite{gu2021unidoc,huang2022layoutlmv3} propose to pre-train multi-modal Transformers for document understanding.
However, these methods still employ only visual information for DLA fine-tuning.
This might lead to degraded performances and generalization ability for DLA models.

To better exploit both visual and textual information for the DLA task, grid-based methods~\cite{yang2017learning,kaplan2021combining,zhang2021vsr} cast text with layout information into $2D$ semantic representations (char-grid~\cite{katti2018chargrid,kaplan2021combining} or sentence-grid~\cite{yang2017learning,zhang2021vsr}) and combine them with visual features,
achieving good results in the DLA task.
Although grid-based methods equip with an additional textual input of grid, only visual supervision is used for the model training in the DLA task.
Since there are no explicit textual objectives to guide the linguistic modeling, 
we consider that the capability of semantic understanding is limited in grid-based models, 
compared with the existing document pre-trained models~\cite{huang2022layoutlmv3}.
Therefore, how to effectively model semantic features based on grid representations is a vital step to improve the performance of DLA.
The differences between existing DLA methods are shown in Table~\ref{tab:intro_model_case}.

As a classic Document AI task, there are many datasets for document layout analysis.
However, the variety of existing DLA datasets is very limited.
The majority of documents are scientific papers, even in the two large-scale DLA datasets PubLayNet~\cite{zhong2019publaynet} and DocBank~\cite{li2020docbank} (in Figure \ref{fig:intro_case} (a) \& (b)), which have significantly accelerated the development of document layout analysis recently.
As shown in Figure~\ref{fig:intro_case}, in real-world scenarios, there are diverse types of documents, not limited to scientific papers, but also including letters, forms, invoices, and so on.
Furthermore, the document layout categories in existing DLA datasets are tailored to scientific-paper-type documents such as titles, paragraphs, and abstracts.
These document layout categories are not diverse enough and thus are not suitable for all types of documents, such as the commonly encountered Key-Value areas in invoices and the line-less list areas in budget sheets.
It is evident that a significant gap exists between the existing DLA datasets and the actual document data in the real world, which hinders the further development of document layout analysis and real-world applications.

In this paper, we present \textbf{VGT}, a two-stream multi-modal \textbf{V}ision \textbf{G}rid \textbf{T}ransformer for document layout analysis,
of which \textbf{G}r\textbf{i}d \textbf{T}ransformer (GiT) is proposed to directly model $2D$ language information.
Specifically, we represent a document as a  $2D$ token-level grid as in the grid-based methods~\cite{yang2017learning,kaplan2021combining,zhang2021vsr} 
and feed the grid into GiT.
For better token-level and segment-level semantic awareness,
we propose two new pre-training objectives for GiT. 
First, inspired by BERT~\cite{devlin2018bert}, the \textbf{M}asked \textbf{G}rid \textbf{L}anguage \textbf{M}odeling (MGLM) task is proposed to learn better token-level semantics for grid features,
which randomly masks some tokens in the $2D$ grid input, and recovers the original text tokens on the document through its $2D$ spacial context.
Second, the \textbf{S}egment \textbf{L}anguage \textbf{M}odeling (SLM) task is proposed to enforce the understanding of segment-level semantics in the grid features,
which aims to align the segment-level semantic representations from GiT with pseudo-features generated by existing language model (\eg, BERT~\cite{bao2021beit} or LayoutLM~\cite{xu2020layoutlm}) via contrastive learning.
Both token-level and segment-level features are obtained from the $2D$ grid features encoded by GiT via RoIAlign~\cite{he2017mask}, according to the coordinates.
Combining image features from \textbf{Vi}sion \textbf{T}ransformer (ViT) further,
VGT can make full use of textual and visual features from GiT and ViT respectively 
and leverage multi-modal information for better document layout analysis, especially in text-related classes.

\begin{table}[t]\centering
\small
\setlength{\tabcolsep}{2pt}
\ra{1.2}
\caption{Comparisons with previous DLA datasets.}
\label{tab:datasets}
\begin{tabular}{cccccc}
\toprule[1pt]
\textbf{Dataset} & \textbf{Type} & \textbf{Category} & \textbf{Labeling} &\textbf{Training} & \textbf{Validation} \\
\midrule
PubLayNet &1 & 5	&XML &335,703	&11,245	 \\
DocBank & 1 & 13	&\LaTeX &400,000	&50,000 \\
D$^4$LA & 12 & 27	&Manual &8,868	&	2,224 \\
\bottomrule[1pt]
\end{tabular}
\end{table}

\begin{figure*}[!htp]\centering
 \includegraphics[width=1.0\textwidth]{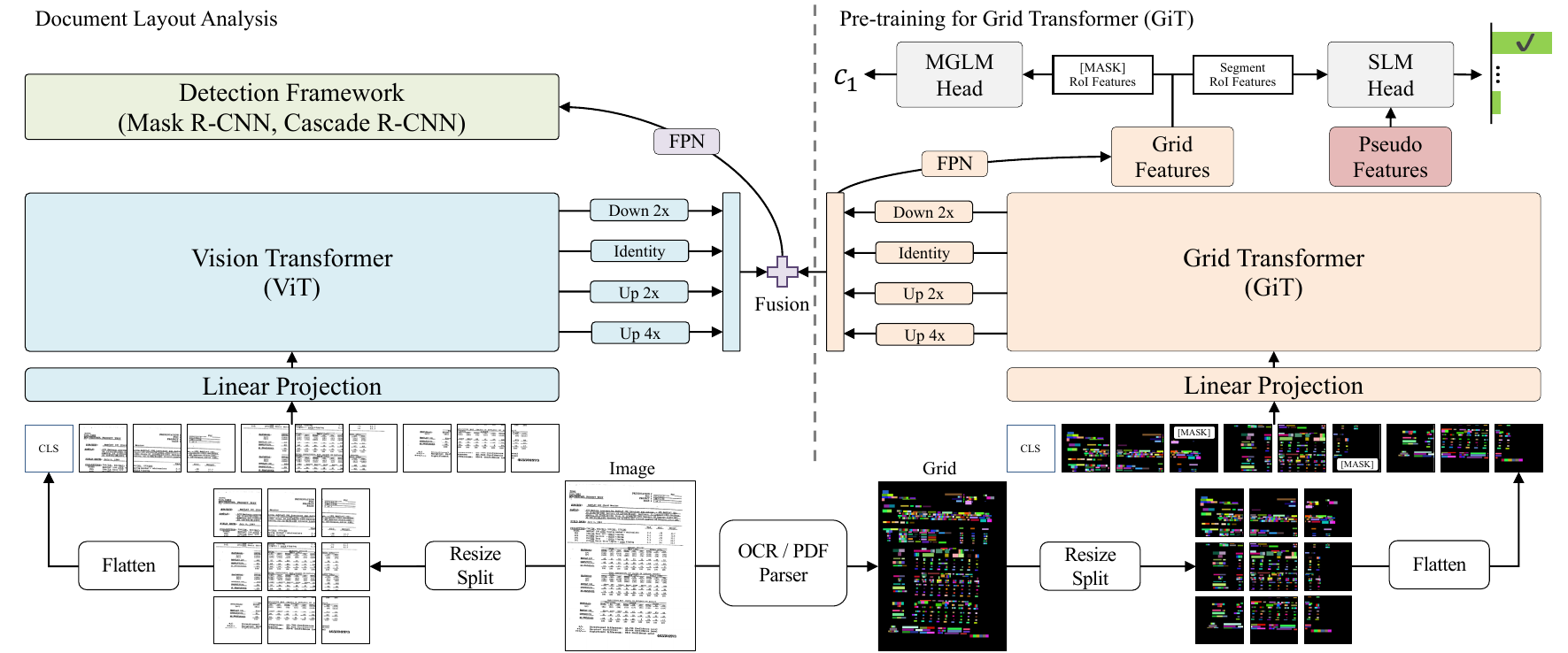}
\caption{The model architecture of Vision Grid Transformer (VGT) with pre-training objectives for the GiT branch.}
 \label{fig:overview}
\end{figure*}

In addition, to facilitate the further advancement of DLA research for real-world applications, we propose the \textbf{D$^4$LA} dataset, which is the most \textbf{D}iverse and \textbf{D}etailed \textbf{D}ataset ever for \textbf{D}ocument \textbf{L}ayout \textbf{A}nalysis. 
The differences from the existing datasets for DLA are listed in Table~\ref{tab:datasets}.
Specifically, D$^4$LA dataset contains $12$ types of documents as shown in Figure~\ref{fig:intro_case}.
We define $27$ document layout categories and \textbf{manually} annotate them.
Some special layout classes are illustrated in Figure~\ref{fig:vis_cats}, which are more challenging and text-related.
Experiment results on DocBank, PubLayNet and D$^4$LA show the SOTA performance of VGT.

The contributions of this work are as follows:

\begin{itemize} \setlength{\itemsep}{0pt}
  \item[1)]  We introduce VGT, a two-stream Vision Grid Transformer for document layout analysis, which can leverage token-level and segment-level semantics in the text grid by two new proposed pre-training tasks: MGLM and SLM.
  To the best of our knowledge, VGT is the first to explore grid pre-training for $2D$ semantic understanding in documents.
  \item[2)]  A new benchmark named D$^4$LA, which is the most diverse and detailed manually-labeled dataset for document layout analysis, is released. It contains $12$ types of documents and $27$ document layout categories.
  \item[3)]  Experimental results on the existing DLA benchmarks (PubLayNet and DocBank) and the proposed D$^4$LA dataset demonstrate that the proposed VGT model outperforms previous state-of-the-arts.
\end{itemize}

\section{Related Works}
Document Layout Analysis (DLA) is a long-term research topic in computer vision~\cite{binmakhashen2019document}.
Previous methods are rule-based approaches~\cite{rule1,rule2} that directly use the image pixels or texture for layout analysis.
Moreover, some machine learning methods~\cite{ml1,ml2} employ low-level visual features for document parsing.
Recent deep learning works~\cite{schreiber2017deepdesrt,CascadeTabNet} consider DLA as a classic visual object detection or segmentation problem and utilize convolutional neural networks (CNN)~\cite{he2016deep} to solve this task~\cite{ren2015faster,liu2016ssd, redmon2018yolov3, he2017mask}.

Self-supervised pre-training techniques have given rise to blossom in Document AI~\cite{xu2020layoutlm,xu2020layoutlmv2,li2021structext,appalaraju2021docformer,luo2022bivldoc,huang2022layoutlmv3,li2022dit,gu2021unidoc,Luo_2023_CVPR}. 
Some document pre-trained models~\cite{huang2022layoutlmv3,li2022dit,gu2021unidoc} have been applied to the DLA task and achieved good performances.
Inspired by BEiT~\cite{bao2021beit}, DiT~\cite{li2022dit} trains a document image transformer for DLA and obtains promising performance, while neglecting the textual information in documents.
Unidoc~\cite{gu2021unidoc} and LayoutLMv3~\cite{huang2022layoutlmv3} model the documents in a unified architecture with the text, vision and layout modalities, but they only use the vision backbone without text embeddings for object detection during fine-tuning the downstream DLA task.
Different from the methods that regard DLA as a vision task, LayoutLM~\cite{xu2020layoutlm} regards the DLA task as a sequence labeling task to explore DLA only in text modality.
The experimental results of LayoutLM show the possibility to use NLP-based methods to process DLA tasks.
However, these methods solely use a single modality for the DLA task, and most of them focus on visual information and  ignore textual information.

To model the documents in multi-modality for the DLA task, like the grid-based models~\cite{katti2018chargrid,denk2019bertgrid,lin2021vibertgrid} in vision information extraction, some works~\cite{yang2017learning,zhang2021vsr} use text and layout information to construct the text grid (text embedding map) and combine it with the visual features for DLA.
Yang \etal~\cite{yang2017learning} build a sentence-level grid that is concatenated with visual features in the model for the DLA task.
VSR~\cite{zhang2021vsr} uses two-stream CNNs where the visual stream and semantic stream take images and text grids (char-level and sentence-level) as inputs, respectively for the DLA task.
However, the text grids in these methods are only as model inputs or extra features, there are no semantic supervisions during DLA task training. Therefore, it is difficult to achieve remarkable semantic understandings.

Previous DLA datasets~\cite{AntonacopoulosBPP09,ClausnerPPA15,yang2017learning} often focus on newspapers, magazines, or technical articles, the size of which is relatively small. 
Recently, the introduction of large-scale DLA datasets such as DocBank~\cite{li2020docbank} and PubLayNet~\cite{zhong2019publaynet} has promoted significant progress in DLA research.
DocBank~\cite{li2020docbank} has 500K documents of scientific papers with $12$ types of layout units.
PubLayNet~\cite{zhong2019publaynet} includes 360k scientific papers with $5$ layout types such as text, title, list, figure, and table.
Since the majority of documents of them are scientific papers,
the variety of document types and layout categories are very limited.
Furthermore, the document layout categories designed for scientific papers in the existing DLA datasets are difficult to transfer to other types of documents in real-world applications.

\section{Vision Grid Transformer}

The overview of Vision Grid Transformer (VGT) is depicted in Figure~\ref{fig:overview}.
VGT employs a Vision Transformer (ViT)~\cite{vit} and a \textbf{G}r\textbf{i}d \textbf{T}ransformer (GiT) to extract visual and textual features respectively, resulting in a two-stream framework.
Additionally,  GiT is pre-trained with the MGLM and SLM objectives to fully explore multi-granularity textual features in a self-supervised and supervised learning manner, respectively.
Finally, the fused multi-modal features generated by the multi-scale fusion module are used for layout detection.

\subsection{Vision Transformer}
\label{sec:Image}
Inspired by ViT~\cite{vit} and DiT~\cite{li2022dit} , 
we directly encode image patches as a sequence of patch embeddings for image representation by linear projection.
In Figure~\ref{fig:overview}, a resized document image is denoted as $\mathbf{I} \in  \mathbb{R}^{H \times W \times C_{I}} $,
where $H$, $W$ and $C_{I}$ are the height, width and channel size, respectively.
Then, $\mathbf{I}$ is split into non-overlapping $P \times P$  patches, and reshaped into a sequence of flattened $2D$ patches $ \mathbf{X}_I \in  \mathbb{R}^{N \times (P^2  C_{I})} $.
We linearly project $ \mathbf{X}_I$ into $D$ dimensions, resulting in a sequence of $N = HW/P^2$ tokens as $ \mathbf{F}_I \in  \mathbb{R}^{N \times D} $.
Following~\cite{vit}, standard learnable $1D$ position embeddings and a $[CLS]$ token are injected.
The resultant image embeddings serve as the input of ViT. 

\subsection{Grid Transformer}
The architecture of GiT is similar to ViT in Section~\ref{sec:Image},
while the input patch embeddings are grid~\cite{katti2018chargrid,denk2019bertgrid}.
Concretely, given a document PDF, PDFPlumber~\footnote{https://github.com/jsvine/pdfplumber} is used to extract words with their bounding boxes.
While for images, we adopt an open-sourced OCR engine.
Then, a tokenizer is used to tokenize the word into sub-word tokens, 
and the width of the word box is equally split for each sub-word.
The complete texts are represented as $\mathcal{D}= \{(c_k, \mathbf{b}_k) | k = 0,\dots,n$ \}, where $c_k$ denotes the $k$-th sub-word token in the page and $\mathbf{b}_k$ is the associated box of $c_k$. 
Finally, the grid input $\mathbf{G} \in \mathbb{R}^{H \times W \times C_{G}} $ is constructed as follows:

\begin{equation}
\mathbf{G}_{i,j}=
\begin{cases}
\mathbf{E}(c_k), & if \: (i,j) \prec \mathbf{b}_k, \\
\mathbf{E}(\text{[PAD]}), & otherwise,
\end{cases}
\end{equation}
where $\prec $ means point $(i,j)$ is located in the box $\mathbf{b}_k$ 
and thus all pixels in $\mathbf{b}_k$ share the same text embedding $\mathbf{E}(c_k)$ of $c_k$.
$\mathbf{E}(\cdot)$ represents an embedding layer, which maps tokens into feature space.
The background pixels with non-text are set as the embedding of a special token [PAD].

As in ViT, $\mathbf{G} $ is split into $P \times P$  patches 
and flattened into a sequence of patches $ \mathbf{X}_G \in  \mathbb{R}^{N \times (P^2 C_{G})} $.
We also utilize linear projection to transcribe  $ \mathbf{X}_G$ into patch embeddings $ \mathbf{F}_G \in  \mathbb{R}^{N \times D} $.
Similarly, $\mathbf{F}_G$ embeddings added with learnable $1D$ position embeddings and a learnable $[CLS]$ token 
are transferred into GiT, generating $2D$ grid features.

\subsection{Pre-Training for Grid Transformer}

\begin{figure}[t]\centering
 \includegraphics[width=0.45\textwidth]{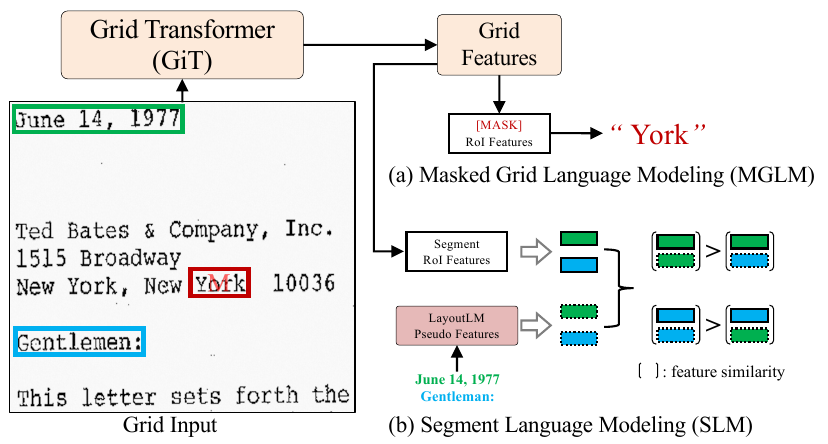}
\caption{Schematic overview of the pre-training for GiT.}
 \label{fig:ptm}
\vspace{-2mm}
\end{figure}

To facilitate the $2D$ understanding of token-level and segment-level semantics in GiT,
we propose Masked Grid Language Modeling (MGLM) and Segment Language Modeling (SLM) objectives for GiT pre-training.
Notably, we decouple visual and language pre-training and \textbf{ONLY} pre-train GiT within grid inputs in VGT.
The reasons are as follows:
(1) Flexibility: different pre-training strategies can be used for ViT pre-training, such as ViT~\cite{vit}, BEiT~\cite{bao2021beit} and DiT\cite{li2022dit}.
(2) Alignment: language information rendered as $2D$ grid features by GiT are naturally well-aligned with image ones in spatial position, and so that the alignment learning in~\cite{xu2020layoutlmv2,huang2022layoutlmv3} is not that necessary.
(3) Efficiency: it can speed up the pre-training process.
The schematic overview of the pre-training for GiT is shown in Figure~\ref{fig:ptm}.

\noindent
\textbf{Masked Grid Language Modeling (MGLM).}
MLM objective~\cite{devlin2018bert} predicts the masked tokens based on the contextual representations, 
of which the output features are readily accessible from a $1D$ sequence via index.
The input and output of GiT, however, are $2D$ feature maps.
We extract the region textual feature with RoIAlign~\cite{maskrcnn} as in RegionCLIP~\cite{regionclip}.
Specifically, we randomly mask some tokens in $\mathbf{G} $ with [MASK] as the input of GiT,
and utilize FPN~\cite{FPN} to generate refined features of GiT as in Figure~\ref{fig:overview}.
Then, the region feature $\mathbf{e}_{c_k}$ of masked token $c_k$
is cropped on the largest feature map (\ie $P_2$ of FPN) by RoIAlign with the box $\mathbf{b}_k$.
The pre-training objective is to maximize the log-likelihood of the correct masked tokens $c_k$ 
based on  $\mathbf{e}_{c_k}$ as
\begin{equation}
L_{MGLM} (\theta) = -\sum_{k=1}^{N_M} \log \; p_{\theta}( c_k \; | \; \mathbf{e}_{c_k} ),
\end{equation}
where $\theta$ is the parameters of GiT and FPN, $N_M$ is the number of masked tokens.

MGLM also differs from variants of MLM used in most of the previous works on layout-aware language modeling.
The key difference between them lies in the way they utilize the $2D$ layout information. In MLM variants (\eg, MVLM in LayoutLM~[35]), the $2D$ position embeddings of text boxes ($[T, 4, C]$) are explicitly computed and added to the embeddings of text sequences ($[T, C]$), whereas in MGLM the $2D$ spatial arrangement is naturally preserved in the $2D$ grid $\mathbf{G}$ ($[H,W,C]$) and explicit $2D$ position embeddings of text are unnecessary.

\noindent
\textbf{Segment Language Modeling (SLM).}
Token-level representation can be efficiently explored vis MGLM task.
However, extremely precise token-level features may not be that crucial in DLA.
Segment-level representation is also essential for object detection.
Thus, the SLM task is proposed to explore the segment-level feature learning of text.
Concretely, we use PDFMiner~\footnote{https://github.com/euske/pdfminer} to extract text lines with bounding boxes as segments.
Then, an existing language model (\eg, BERT or LayoutLM) is used to generate the feature $\mathbf{e}_{l_i}^{*} $ of segment $l_i$ as pseudo-target.
The segment feature $\mathbf{e}_{l_i} $ of $l_i$ is produced by RoIAlign with its line box.
Given the aligned segment-target feature pairs $\{(\mathbf{e}_{l_i}, \mathbf{e}_{l_i}^{*})\}$, contrastive loss~\cite{regionclip} is used for SLM task, which is computed as
\begin{equation}
\begin{split}
& p_{\theta}( \mathbf{e}_{l_i}, \mathbf{e}_{l_i}^{*} )=\dfrac{\exp(\mathbf{e}_{l_i} \cdot \mathbf{e}_{l_i}^{*}/\tau)}{\exp(\mathbf{e}_{l_i} \cdot \mathbf{e}_{l_i}^{*}/\tau) + \sum_{k\in{\mathcal{N}_{l_i}}}\exp(\mathbf{e}_{l_i} \cdot  \mathbf{e}_{l_k}^{*}/\tau)}  \\
& L_{SLM} (\theta) = -\dfrac{1}{N_S} \sum_{i=1} \log \; p_{\theta}( \mathbf{e}_{l_i}, \mathbf{e}_{l_i}^{*} ). \nonumber
\end{split}
\end{equation}
Here, $\cdot$ represents the cosine similarity between segment feature $\mathbf{e}_{l_i}$ from FPN and the pseudo-target feature $\mathbf{e}_{l_i}^{*}$ generated by language model.
$\mathcal{N}_{l_i}$ represents a set of negative samples for segment $l_i$,
and $\tau$ is a predefined temperature. 
We sample $N_S$ segments on one page.
Finally, $L_{MGLM}$ and $L_{SLM}$ are equally used for GiT pre-training.

\subsection{Multi-Scale Multi-Modal Feature Fusion}
FPN~\cite{FPN} framework is widely used to extract multi-scale features in object detection~\cite{maskrcnn}.
To adapt the single-scale ViT to the multi-scale framework,
we use $4$ resolution-modifying modules at different transformer blocks, following~\cite{li2022dit}.
In this way, we obtain multi-scale features of ViT and GiT, denoted as 
$\{ \mathbf{V}_i \in  \mathbb{R}^{ H/2^i \times W/2^i  \times D} | i = 2\ldots,5\}$ and $\{ \mathbf{S}_i \in  \mathbb{R}^{ H/2^i \times W/2^i  \times D} | i = 2\ldots,5\}$, respectively.
Then, we fuse the features at each scale $i$ respectively as
\begin{equation}
\mathbf{Z}_i = \mathbf{V}_i \oplus \mathbf{S}_i, \quad  i = 2\ldots,5,
\end{equation}
where $\oplus$ represents an element-wise sum function in our implements.
Then, we employ FPN to refine pyramid features $\{ \mathbf{Z}_i \in  \mathbb{R}^{ H/2^i \times W/2^i  \times D}| i = 2\ldots,5\}$  further.
Finally, we can extract RoI features from different levels of the feature pyramid according to their scales for later object detection.

\section{D$^4$LA Dataset}
\begin{table*}[t]\centering
\setlength{\tabcolsep}{2pt}
\ra{1.2}
\caption{Statistics of different document types of training and validation sets in the D$^4$LA dataset.}
\label{tab:d3oc}
\begin{tabular}{|l|l|l|l|l|l|l|}
\hline
Category &
  \begin{tabular}[c]{@{}l@{}}\textbf{Budget}\\ 845 / 212\end{tabular} &
  \begin{tabular}[c]{@{}l@{}}\textbf{Email}\\ 780 / 195\end{tabular} &
  \begin{tabular}[c]{@{}l@{}}\textbf{Form}\\ 650 / 163\end{tabular} &
  \begin{tabular}[c]{@{}l@{}}\textbf{Invoice}\\ 574 / 144\end{tabular} &
  \begin{tabular}[c]{@{}l@{}}\textbf{Letter}\\ 793 / 199\end{tabular} &
  \begin{tabular}[c]{@{}l@{}}\textbf{Memo}\\ 817 / 205\end{tabular} \\ \cline{2-7} 
Training / Validation &
  \begin{tabular}[c]{@{}l@{}}\textbf{News article}\\ 682 / 171\end{tabular} &
  \begin{tabular}[c]{@{}l@{}}\textbf{Presentation}\\ 721 / 181\end{tabular} &
  \begin{tabular}[c]{@{}l@{}}\textbf{Resume}\\ 854 / 214\end{tabular} &
  \begin{tabular}[c]{@{}l@{}}\textbf{Scientific publication}\\ 760 / 190\end{tabular} &
  \begin{tabular}[c]{@{}l@{}}\textbf{Scientific report}\\ 616 / 155\end{tabular} &
  \begin{tabular}[c]{@{}l@{}}\textbf{Specification}\\ 776  / 195 \end{tabular} \\ \hline
\end{tabular}
\end{table*}

\begin{table*}[t]\centering
\small
\setlength{\tabcolsep}{2pt}
\ra{1.2}
\caption{Statistics of different layout categories of training and validation sets in the D$^4$LA dataset ( \#instances / percentage \%).}
\label{tab:V12_cats}
\begin{tabular}{|l|l|l|l|l|l|l|l|l|l|}
\hline
Category & \textbf{DocTitle} & \textbf{ListText} & \textbf{LetterHead} & \textbf{Question} & \textbf{RegionList}  &\textbf{TableName}  &\textbf{FigureName}  &\textbf{Footer} &\textbf{Number} \\
\cline{2-10}
Training & 7391 / 6.30 & 4581 / 3.90 & 570 / 0.49	& 113 / 0.10  & 3741 / 3.19 & 640  / 0.55 & 295 / 0.25  & 642 /  0.55 & 7289 / 6.21 \\
Validation & 1893 / 6.41 & 1137 / 3.85 & 127 / 0.43	&  56 / 0.19 & 891 / 3.02 &  178 / 0.60 & 85 / 0.29  &  170 / 0.58 &  1833 / 6.21 \\
\hline
Total & \textbf{ParaTitle} & \textbf{RegionTitle} &\textbf{LetterDear}   &\textbf{OtherText}  & \textbf{Abstract} &\textbf{Table} & \textbf{Equation}  & \textbf{PageHeader}  & \textbf{Catalog} \\
\cline{2-10}
Training  & 4962 / 4.23 & 5469 / 4.66  &   871 / 0.74 &  15229 / 12.98 & 807 / 0.69  & 2733 / 2.33  &  54 / 0.05 & 3941 /  3.36 & 21 / 0.02   \\
117322 &   1333 / 4.51 &  1352 / 4.58  &  228 / 0.77 &  3703 / 12.54  &  200 / 0.68  &  656 / 2.22  &   20 / 0.07 & 933 / 3.16  & 14 / 0.05 \\
\hline
Total & \textbf{ParaText} & \textbf{Date} &\textbf{LetterSign}   &\textbf{RegionKV}  & \textbf{Author} &\textbf{Figure} & \textbf{Reference}  & \textbf{PageFooter}  & \textbf{PageNumber} \\
\cline{2-10}
Validation & 32328 / 27.55  & 3148 / 2.68 & 738 / 0.63 	& 12322 / 10.50   & 1384 / 1.18  &  2201 / 1.88  & 574 / 0.49  & 3164 / 2.70 &  2114 / 1.80  \\
29524 & 8400 / 28.45 &  786 / 2.66 & 175 / 0.59	& 2947 / 9.98   & 371 / 1.26  & 592 /2.01 &  148 / 0.50 & 797 / 2.70 & 499 / 1.69  \\
\hline
\end{tabular}
\end{table*}

In this section, we introduce the proposed D$^4$LA dataset.

\noindent
\textbf{Document Description.}
The images of D$^4$LA are from RVL-CDIP~\cite{rvl}, which is a large-scale document classification dataset in $16$ classes.
We choose $12$ types documents with rich layouts from it,
and sample about $1,000$ images of each type for \textbf{manual} annotation.
The noisy, handwritten, artistic or less text images are filtered.
The OCR results are from IIT-CDIP~\cite{iit}.
The statistics on different document types of D$^4$LA dataset are listed in Table~\ref{tab:d3oc}.

\noindent
\textbf{Category Description.}
We define detailed $27$ layout categories for real-world applications, 
\ie, DocTitle, ListText, LetterHead, Question, RegionList, TableName, FigureName, Footer, Number,
ParaTitle, RegionTitle, LetterDear, OtherText, Abstract, Table, Equation, PageHeader, Catalog, ParaText, Date, LetterSign, RegionKV, Author, Figure, Reference, PageFooter, and PageNumber.
For example, we define $2$ region categories for information extraction task, \ie,
RegionKV is a region that contains Key-Value pairs and RegionList for wireless form as in Figure~\ref{fig:vis_cats}.
The statistics of each category of D$^4$LA are listed in Table~\ref{tab:V12_cats}.
More detailed descriptions and examples can be found in supplementary materials.

\noindent
\textbf{Characteristics.}
Documents in existing large-scale DLA datasets~\cite{zhong2019publaynet,li2020docbank} are mainly scientific papers, where documents in real-world scenarios are not well represented.
In contrast, D$^4$LA includes $12$ diverse document types and $27$ detailed categories.
The variety of types and categories is substantially enhanced, closer to the use of real-world applications.
Moreover, the image quality of D$^4$LA  may be poor, \ie, scanning copies are noisy, skew or blurry.
The increased diversities and low-quality scanned document images constitute a more challenging benchmark for DLA.

\section{Experiments}
\subsection{Implementation Details}
\noindent
\textbf{Model Configuration.}  
VGT is built upon two ViT-Base models, which adopt a $12$-layer Transformer encoder 
with $12$-head self-attention, $D=768$ hidden size and $3,072$ intermediate size of MLP~\cite{vit}.
The patch size $P$ is $16$ as in~\cite{vit} for both ViT and GiT.
For grid construction, WordPiece tokenizer~\cite{devlin2018bert} is used to tokenize words.
We initialize the text embeddings of GiT with those of LayoutLM~\cite{xu2020layoutlm}
and reduce the embedding size $768$ to $C_G=64$ for memory constraints.
$\mathbf{G}$ has the same shape as the original image.

\noindent
\textbf{Model Pre-Training.} 
For ViT pre-training, we directly use the weights of the DiT-base model~\cite{li2022dit}.
For GiT, we also initial it with the weights of the DiT-base model and perform pre-training for GiT on a subset of IIT-CDIP~\cite{iit} dataset with about $4$ million images via MGLM and SLM tasks.
Specifically, we randomly set some tokens as [MASK] tokens, and recover the masked tokens on MGLM task as in BERT~\cite{devlin2018bert}.
For SLM task, we randomly select $N_S=64$ segments of one page
and employ LayoutLM~\cite{xu2020layoutlm} to generate $\mathbf{e}^*$ as pseudo-targets.
We use Adam optimizer with $96$ batch size for 150,000 steps.
We use a learning rate $5e-4$ and linearly warm up $2\%$ first steps.
The image shape is $768 \times 768$ and $\tau$ is $0.01$.

\noindent
\textbf{Model Fine-Tuning.} 
We treat layout analysis as object detection
and employ VGT as the feature backbone in the Cascade R-CNN~\cite{cascade} detector with FPN~\cite{FPN},
which is implemented based on the Detectron2~\cite{Detectron2}.
AdamW optimizer with 1,000 warm-up steps is used, and the learning rate is $2e-4$.
We train VGT for 200,000 steps on DocBank and 120,000 steps on PubLayNet with $24$ batch size.
Since D$^4$LA is relatively small, we train VGT for 10,000 steps on it with $12$ batch size.
The other settings of Cascade R-CNN are the same with DiT~\cite{li2022dit}. 

\subsection{Datasets}
Besides the proposed D$^4$LA dataset, two benchmarks for document layout analysis are used for evaluation.
For the visual object detection task,
we use the category-wise and overall mean average precision (mAP) @IOU[0.50:0.95] of bounding boxes as the evaluation metric.

\noindent
\textbf{PubLayNet}~\cite{zhong2019publaynet} contains 360K research PDFs for DLA released by IBM.
The annotation is in object detection format with bounding boxes and polygonal segmentation in $5$ layout categories (Text, Title, List, Figure, and Table). 
Following~\cite{gu2021unidoc,li2022dit,zhong2019publaynet}, we train models on the training split (335,703) and evaluate on the validation split (11,245).

\noindent
\textbf{DocBank}~\cite{li2020docbank} includes 500K document pages with fine-grained token-level
annotations released by Microsoft.
Moreover, region-level annotations in $13$ layout categories (Abstract, Author, Caption, Equation, Figure, Footer,
List, Paragraph, Reference, Section, Table and Title) are proposed for object detection.
We train models on the training split (400K), and evaluate on the validation split (5K)~\cite{li2020docbank}.

Since both PubLayNet and DocBank datasets are relatively large, 
we construct two sub-datasets PubLayNet2K and DocBank2K 
that sample $2,000$ images for training and $2,000$ images for validation respectively, 
to quickly verify the effects of different modules of VGT in the early experiments. 
We train VGT for $10,000$ steps on them.

\subsection{Discussions on GiT}
We deeply study the effectivenesses of GiT in Table~\ref{tab:PTMabla}, 
where (a) is a single-stream baseline with only ViT.

\noindent
\textbf{Effectiveness of Only GiT.} 
We can directly employ GiT for layout detection since the grid input of GiT naturally contains the fine-grained layout and textual information. To disentangle the effect of layout and text, 
we use GiT as the feature backbone and set all the sub-word tokens as the [UNK] token in (c), 
where no textual messages are used.
We then adopt the original tokens as input in (d) to verify the effectiveness of the text.
Both (c) and (d) are not pre-trained, and we pre-train GiT with MGLM and SLM objectives in (e). We observe that only layout information can produce rough detection results,
adding text brings improvements, and pre-training objectives can further exploit the ability of GiT.
Notably, DocBank2K contains more linguistic categories than PubLayNet2K,
such as ``Date'', ``Author'' and so on.
Therefore, the improvement on DocBank2K is more remarkable than that on PubLayNet2K in (d) and (e).
These results demonstrate that sub-word layout information can be directly used for layout analysis, 
textual cues of the grid can indeed facilitate layout analysis, 
and a well pre-trained Grid Transformer for grid inputs is indispensable.

\begin{table}[t]\centering
\footnotesize
\setlength{\tabcolsep}{1pt}
\ra{1.1}
\caption{Effect of different modules of GiT on PubLayNet2K and DocBank2K.}
\label{tab:PTMabla}
\begin{tabular}{|c|c|c|c|c|c|c|c|}
\hline
\multirow{2}{*}{\textbf{Tag}}  & \textbf{Image} & \textbf{Grid} & \textbf{Grid} & \multirow{2}{*}{\textbf{PTM}}  &\textbf{PubLayNet} & \textbf{DocBank} \\
 & \textbf{Backbone} & \textbf{Backbone} & \textbf{Embedding} &  & \textbf{2K} &  \textbf{2K} \\
\hline
(a) & ViT 	& - &- & - &86.92  &59.61 \\
(b)  & ViT  	& ViT	&Image  &\ding{55}	  &86.98 &58.57 \\
\hline
(c)  & - 	& GiT &[UNK] & \ding{55} &64.12  &40.56 \\
(d)  & - 	& GiT &LayoutLM & \ding{55}  &65.88  &49.15 \\
(e)  & - 	& GiT &LayoutLM & \ding{51}  &74.96  &55.46 \\
\hline
(f)  & ResNeXt-101 	& - &- &- &83.54  &57.04 \\
(g)  & ResNeXt-101  	& GiT &LayoutLM &\ding{51}  &85.58  &63.05 \\
\hline
(h)  & ViT  	& GiT	&BERT  &\ding{55}  &87.97  &63.97 \\
(i)  & ViT  	& GiT	&LayoutLM  &\ding{55}	  &87.76  &64.01 \\
(j)  & ViT  	& GiT	&LayoutLM  &\ding{51} 	 &\textbf{88.44}  &\textbf{65.94} \\
\hline
\end{tabular}
\end{table}

\noindent
\textbf{Effectiveness of VGT.} 
We compare the performance between different word embeddings, 
\ie BERT~\cite{devlin2018bert} in (h) and LayoutLM~\cite{xu2020layoutlm} in (i).
The results show that the VGT with both word embeddings can lead to better performance than (a).
Since LayoutLM is pre-trained on documents and possesses the capability of layout modeling,
we use the embeddings of LayoutLM in the following experiments.
Moreover, we introduce a pre-training mechanism in (j), resulting in significant improvements over (i).
These results verify the effectiveness of VGT and the pre-training for GiT.

\noindent
\textbf{Compatibility of GiT.} 
Due to the decoupling framework of VGT,
we can perform solely pre-training for GiT and further integrate GiT with not only ViT but also CNNs.
We train a Cascade R-CNN with ResNeXt-101~\cite{resnext} backbone and FPN in (f) as a baseline.
Typically, we construct a hybrid model (g) with ResNeXt-101 and the pre-trained GiT.
Clearly, the results of (g) can surpass that of (f), 
demonstrating the good compatibility of the pre-trained GiT.

\noindent
\textbf{Effect of Parameters.} 
Using the two-stream framework inevitably leads to an increase in the number of model parameters.
We conduct experiments to analyze the effect of model parameters.
In (b), we replace GiT with one ViT,  and thus simply construct a two-stream ViT framework with two image inputs.
Comparing (b) with (a), introducing more parameters with double image inputs can not enhance the capability of the model.
Referring to (g), (h), (i) and (j), GiT models the layout and textual information as a supplement to image information,
resulting in obvious improvements.

\subsection{Ablation Study of Pre-Training Objectives}
We investigate the effect of the proposed pre-training objectives on more text-related DocBank2K in Table~\ref{tab:PTM}.
Model (a) is the baseline of VGT without pre-training.
We pre-train (b) with only MGLM  and (c) with only SLM.
Then, both MGLM and SLM objectives are utilized for (d).
All models are trained on $0.5$ million images with $525,000$ steps for experimental efficiency.
Model (b) obtains better performance than (a)
indicating that predicting masked tokens in MLM objective~\cite{devlin2018bert} makes sense in $2D$ textual space.
Notably, model (c) attains a small improvement over (b).
We speculate that the textual features of segments are more suitable for the layout detection task, 
and thus segment-level SLM works better than token-level MGLM.
Model (d) with MGLM and SLM can achieve the best results.

\begin{table}[t]\centering
\setlength{\tabcolsep}{6pt}
\ra{1.2}
\caption{Ablation study of pre-training objectives.}
\label{tab:PTM}
\begin{tabular}{|c|c|c|c|}
\hline
\textbf{Tag}  & \textbf{MGLM}  & \textbf{SLM}  & \textbf{DocBank2K} \\
\hline
(a)  &- 	&-   &64.010 \\
(b)   &\ding{51} 	&\ding{55} &64.539 \\
(c)   &\ding{55} 	&\ding{51}  &65.112 \\
(d)   &\ding{51} 	&\ding{51}  &65.167  \\
\hline
\end{tabular}
\end{table}

\begin{table}[t]\centering
\small
\setlength{\tabcolsep}{3pt}
\ra{1.2}
\caption{Document Layout Detection mAP @ IOU [0.50:0.95] on PubLayNet validation set.}
\label{tab:pub}
\begin{tabular}{|c|c|c|c|c|c|c|}
\hline
\textbf{Models}  & \textbf{Text} & \textbf{Title} & \textbf{List} & \textbf{Table}  &\textbf{Figure}  & \textbf{mAP} \\
\hline
ResNeXt-101  &93.0 &86.2 &94.0	&97.6 &96.8 &93.5 \\
DiT-Base &94.4 &88.9 &94.8	&97.6 &96.9 &94.5 \\
LayoutLMv3-Base &94.5 &90.6 	&95.5 &97.9  &97.0 &95.1 \\
VSR  &\textbf{96.7} &93.1 	&94.7 &97.4 &96.4 &95.7 \\
\hline
VGT (ours)  &95.0 &\textbf{93.9}	&\textbf{96.8} &\textbf{98.1} &\textbf{97.1} &\textbf{96.2}\\
\hline
\end{tabular}
\end{table}

\begin{table*}[t]\centering
\small
\setlength{\tabcolsep}{1pt}
\ra{1.2}
\caption{Document Layout Detection mAP @ IOU [0.50:0.95] on DocBank validation set.}
\label{tab:docbank}
\begin{tabular}{|c|c|c|c|c|c|c|c|c|c|c|c|c|c|c|}
\hline
\textbf{Models} & \textbf{Abstract} & \textbf{Author} & \textbf{Caption} & \textbf{Date} & \textbf{Equation} & \textbf{Figure}  &\textbf{Footer} &\textbf{List} & \textbf{Paragraph} &\textbf{Reference}   &\textbf{Section}  & \textbf{Table} &\textbf{Title} & \textbf{mAP} \\
\hline
ResNeXt-101  &89.7 &72.6 &82.3	&69.7 &76.4 &73.6 &78.2 &78.3 &66.2 &81.7 &75.9 &77.3 & 84.1 & 77.4 \\
DiT-Base   &91.1  &75.4  &83.1  &73.4  &77.8  &75.7  &80.2  &82.7  &67.3  &83.8  &77.0 &80.8  &86.8 & 79.6  \\
LayoutLMv3-Base & 90.5  &73.6  &81.2  &73.5  &76.0  &74.4  &78.1 &80.7 &65.8  &82.8  &76.6 &78.6 &86.3 &78.3 \\
\hline
VGT (ours) &\textbf{92.4}   &\textbf{79.9} &\textbf{88.8}  &\textbf{79.1}  &\textbf{86.7}  &\textbf{76.6}  &\textbf{84.8}  &\textbf{88.6}  &\textbf{75.8}  &\textbf{85.6}  &\textbf{81.5} &\textbf{83.9}  &\textbf{89.8}  &\textbf{84.1} \\
\hline
\end{tabular}
\end{table*}

\begin{table*}[t]\centering
\footnotesize
\setlength{\tabcolsep}{2.5pt}
\ra{1.2}
\caption{Document Layout Detection mAP @ IOU [0.50:0.95] on D$^4$LA validation set.}
\label{tab:V12}
\begin{tabular}{|c|c|c|c|c|c|c|c|c|c|c|}
\hline
\textbf{Models}  & \textbf{DocTitle} & \textbf{ListText} & \textbf{LetterHead} & \textbf{Question} & \textbf{RegionList}  &\textbf{TableName}  &\textbf{FigureName}  &\textbf{Footer} &\textbf{Number} & \\
\hline
ResNeXt-101  &70.6  &71.0  &\textbf{82.8} &48.4  &76.1  &66.0  &45.9   &76.2  &83.0  &\\
DiT-Base &\textbf{73.1}  &70.6  &82.2 	&55.0  &80.1  &68.4  &\textbf{51.8}   &\textbf{81.2}  &83.2  & \\
LayoutLMv3-Base &66.8  &56.5  &78.5  &39.3  &72.1  &64.3  &32.1   &72.2  &82.1  &\\
VGT (ours)&72.6  &\textbf{71.3}  &82.3	&\textbf{63.9}  &\textbf{80.2}  &\textbf{68.4}  &46.6  &79.7  &\textbf{83.2}  &\\
\hline
\textbf{Models}  &\textbf{ParaTitle} & \textbf{RegionTitle} &\textbf{LetterDear}   &\textbf{OtherText}  & \textbf{Abstract} &\textbf{Table} & \textbf{Equation}  & \textbf{PageHeader}  & \textbf{Catalog} &\\
\hline
ResNeXt-101 &60.3  &63.8  &73.4  &56.4  &65.7  &\textbf{86.3}  &11.5   &53.7  &32.0 & \\
DiT-Base &\textbf{63.2}  &\textbf{67.5}  &74.5 	&59.2  &73.8  &86.2  &9.2  &56.5  &44.8 &  \\
LayoutLMv3-Base &55.6  &59.5  &70.8 &50.8  &68.2  &80.6  &7.3   &53.1  &37.3 & \\
VGT (ours) &63.0  &67.2  &\textbf{76.7} &\textbf{60.0}  &\textbf{80.4}  &86.0  &\textbf{19.9}   &\textbf{56.9}  &\textbf{40.9} & \\
\hline
\textbf{Models}  &\textbf{ParaText} & \textbf{Date} &\textbf{LetterSign}   &\textbf{RegionKV}  & \textbf{Author} &\textbf{Figure} & \textbf{Reference}  & \textbf{PageFooter}  & \textbf{PageNumber} & \textbf{mAP} \\
\hline
ResNeXt-101  &85.2  &68.4  &69.3 	&68.2  &62.6  &76.7  &83.4   &62.2  &57.9  & 65.1 \\
DiT-Base  &\textbf{86.4}  &69.7  &71.6 	&68.8  &66.0  &\textbf{77.2} &83.4  &65.5  &58.3 & 67.7\\
LayoutLMv3-Base  &81.6  &62.5  &60.4 	&59.4  &59.3  &72.2  &74.9   &62.1  &52.8  & 60.5 \\
VGT (ours) &86.2  &\textbf{71.3}  &\textbf{75.5 }	&\textbf{70.1 } &\textbf{67.6}  &76.7  &\textbf{85.6}   &\textbf{66.5}  &\textbf{58.7}  &  \textbf{68.8}  \\
\hline
\end{tabular}
\end{table*}

\subsection{Comparison with State-of-the-arts}
We evaluate the performance of VGT on three datasets, namely PubLayNet, DocBank, and our proposed D$^4$LA.

\begin{figure*}[!htp]\centering
 \includegraphics[width=0.95\textwidth]{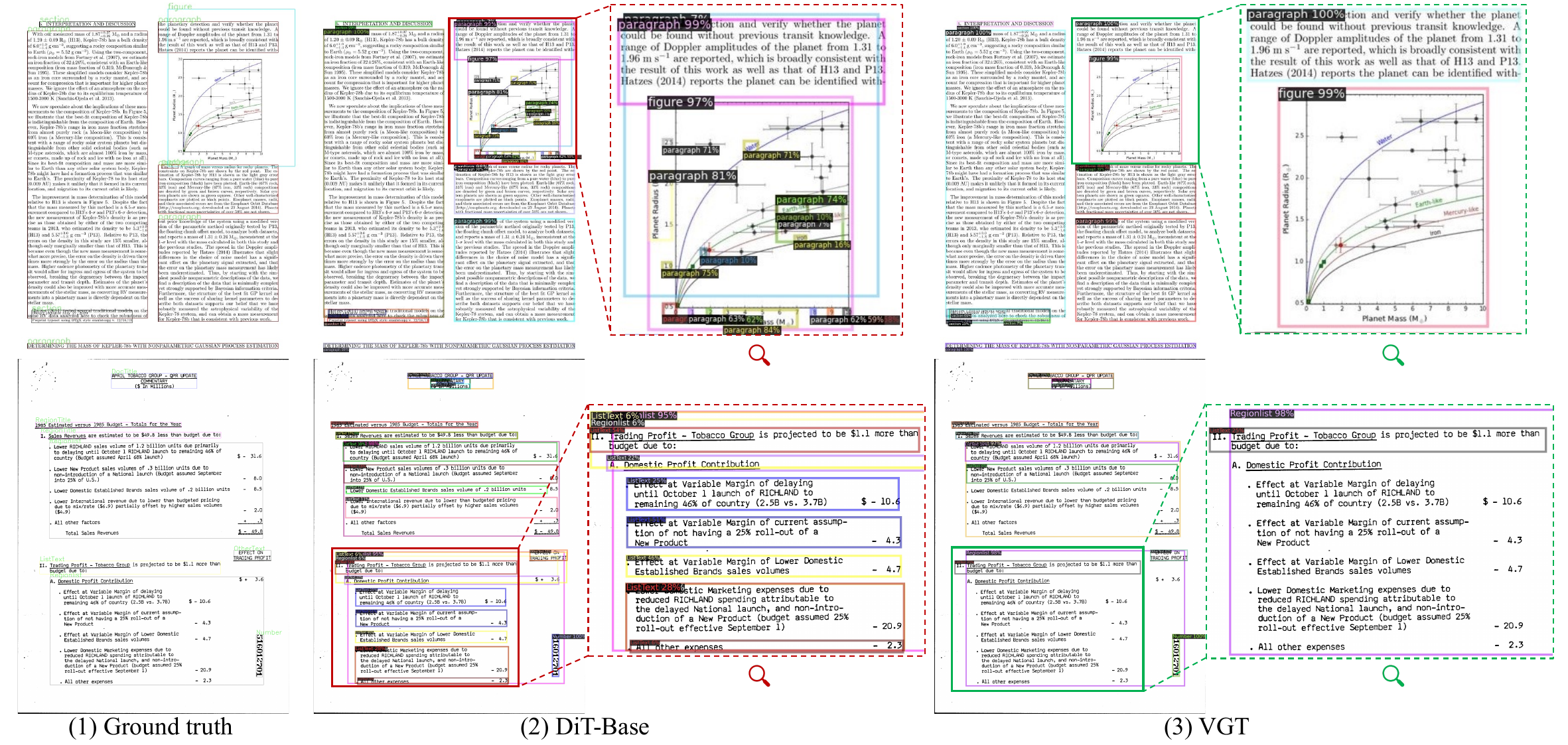}
\caption{Qualitative comparison between DiT-Base and VGT on DocBank (1st row)  and D$^4$LA (2nd row). Best viewed in color.}
 \label{fig:vis_list}
\end{figure*}

\noindent
\textbf{PubLayNet.}  The results of document layout detection on PubLayNet are reported in Table~\ref{tab:pub}.
Generally, PubLayNet contains $5$ relatively simple layout categories,
of which the visual information may be sufficient for layout detection.
Thus, all of the methods can obtain promising mAPs ($>90\%$).
The results of DiT-Base are better than that of ResNeXt-101, 
showing the powerful capability of pre-trained ViT.
LayoutLMv3 pre-trains multi-modal Transformer with unified text and image masking objectives.
We feed only image tokens into LayoutLMv3 without text embeddings as in~\cite{huang2022layoutlmv3}.
LayoutLMv3 achieves better results than pure visual methods (DiT-Base and ResNeXt-101), especially in ``Title'' and ``List'' classes.
Grid-based VSR~\cite{zhang2021vsr} method presents better results in ``Title'' class,
showing the effectiveness of grid for text-related classes.
VGT achieves the best average mAPs and presents a remarkable improvement in ``Title'' and ``List'' classes over VSR.
We attribute this improvement to the textual modeling of GiT and the pre-training objectives.

\noindent
\textbf{DocBank.} 
We measure the mAP performance on DocBank and list the results in Table~\ref{tab:docbank}.
The methods in Table~\ref{tab:docbank} are implemented by the original codes, and the Cascade R-CNN is used for layout detection.
Since DocBank provides more detailed layout categories than PubLayNet,
the mAPs of the text-related categories of DocBank might reflect the ability of textual modeling.
Clearly, the performance of DiT-Base is still better than that of ResNeXt-101, 
showing the superiority of ViT backbone again.
LayoutLMv3 obtains a little worse result than DiT-Base.
Since LayoutLMv3 is pre-trained with text but no explicit text embeddings are used in DLA task,
we conjecture that the text information may be insufficient for detailed detection on DocBank.
VGT obtains the best mAPs of text-related categories and exhibits substantial improvement over other methods, such as ``Caption'', ``Date'', ``Equation'', ``List'' and ``Paragraph'', verifying the effectiveness of VGT.

\noindent
\textbf{D$^4$LA.} 
Due to the more diverse document types and detailed categories of D$^4$LA,
the detection results reported in Table~\ref{tab:V12} are relatively lower, compared with the mAPs of PubLayNet and DocBank.
It reveals that the existing methods do not work well on real-world documents.
Similarly, the DiT-Base backbone achieves better performance than ResNeXt-101 in D$^4$LA,
due to the well-designed image pre-training objective on documents.
LayoutLMv3 obtains worse results than DiT-Base, especially in the text-related classes.
VGT achieves the best results in most categories. 
The significant improvements on text-related categories (\eg, $6.6\%$ on ``Abstract'' over DiT-Base)
verify the superiority of VGT on textual modeling in $2D$ fashion.

\subsection{Visualization Cases}
We illustrate the detection results of DiT-Base and VGT on samples from DocBank and  D$^4$LA in Figure~\ref{fig:vis_list}. For the sample of DocBank, the text in the chart is misidentified as ``Paragraph'' in DiT-Base, while VGT removes the predictions of them and produces a precise box of ``Figure''.
This is because there is no text in the chart for grid construction, beneficial to false positive reduction. For the sample of D$^4$LA, the predictions of ``ListText'' are drastically reduced.
Visually, these regions alone are indeed like ``ListText'' regions.
However, they constitute a ``RegionList'' from the contextual semantics.
These qualitative results demonstrate the ability of $2D$ language modeling in VGT.

\section{Limitations}
Due to the two-stream framework, VGT contains 243M parameters, 
relatively larger than DiT-Base (138M), and LayoutLMv3 (138M).
The inference time of VGT (460ms) is relatively longer than DiT-Base (210 ms) and LayoutLMv3 (270 ms).
Thus, a more lightweight and efficient architecture will be our future work.
Moreover, since VGT is an image-centric method, we will extend VGT to text-centric tasks, such as information extraction in the future.

\section{Conclusion}
In this paper, we present VGT, a two-stream Vision Grid Transformer for document layout analysis.
VGT is an image-centric method, that is more compatible with object detection.
The Grid Transformer of VGT is pre-trained by MGLM and SLM objectives
for $2D$ token-level and segment-level semantic understanding.
In addition, we propose a new dataset D$^4$LA, which is the most diverse and detailed manually-annotated benchmark ever for document layout analysis.
Experimental results show that VGT achieves state-of-the-art results on PubLayNet, DocBank and the proposed D$^4$LA.

{\small
\bibliographystyle{ieee_fullname}
\bibliography{egbib}
}

\clearpage
\appendix

\section*{Appendix}
\section{Annotations of D$^4$LA Dataset}
It is time-consuming and labor-intensive to manually annotate the images of various document types with complex layout categories.
We employ about $5$ full-time annotators to annotate these complex document images for about $1.5$ months.
The definition of categories and the guideline of annotations are carefully designed and can be basically applied to other types of documents.
The layout annotations of the bounding boxes in our D$^4$LA dataset are in standard MSCOCO format for the classic detection task.

For the OCR results of document images in our D$^4$LA dataset, we first map the images of D$^4$LA into RVL-CDIP dataset, 
and further, map them into IIT-CDIP dataset which is the superset of RVL-CDIP and provides the text contents and bounding boxes for each word.
The original images, the OCR results of them and the manual layout annotations will be made publicly available.

\section{Detailed Layout Categories in D$^4$LA}
We describe the definitions of the different layout categories of the proposed D$^4$LA dataset.
We simply introduce the common categories of scientific papers which is similar to those of DocBank.
For some special layout categories of D$^4$LA, we illustrate them in detail.

\subsection{Common Categories in Scientific Papers} 
Documents of the existing large-scale DLA datasets are mainly scientific papers.
The layout categories are defined especially for scientific papers.
These common categories are not only included in D$^4$LA
but also more detailed.
We introduce these common categories as follows:

\noindent
\textbf{DocTitle} is the title of the document that is similar to \textit{Title} of papers in DocBank. 
While, in other types of documents, we define the text at the head of the document as ``DocTitle'',
that is commonly bold or with underlines.

\noindent
\textbf{ListText} is a paragraph with bullet or enumeration symbols,
which is different from \textit{List} in PubLayNet and DocBank.
Specifically,  \textit{List}  is a region where all instances of ``ListText''  are grouped together into one ``List'' object block.
While ``ListText'' is an individual object instance that is one of the paragraphs of \textit{List} region.
The definition of ``ListText'' is more suitable for other types of documents, 
since ``ListText'' instances are often mixed with other text.

\begin{figure}[t]\centering
 \includegraphics[width=0.48\textwidth]{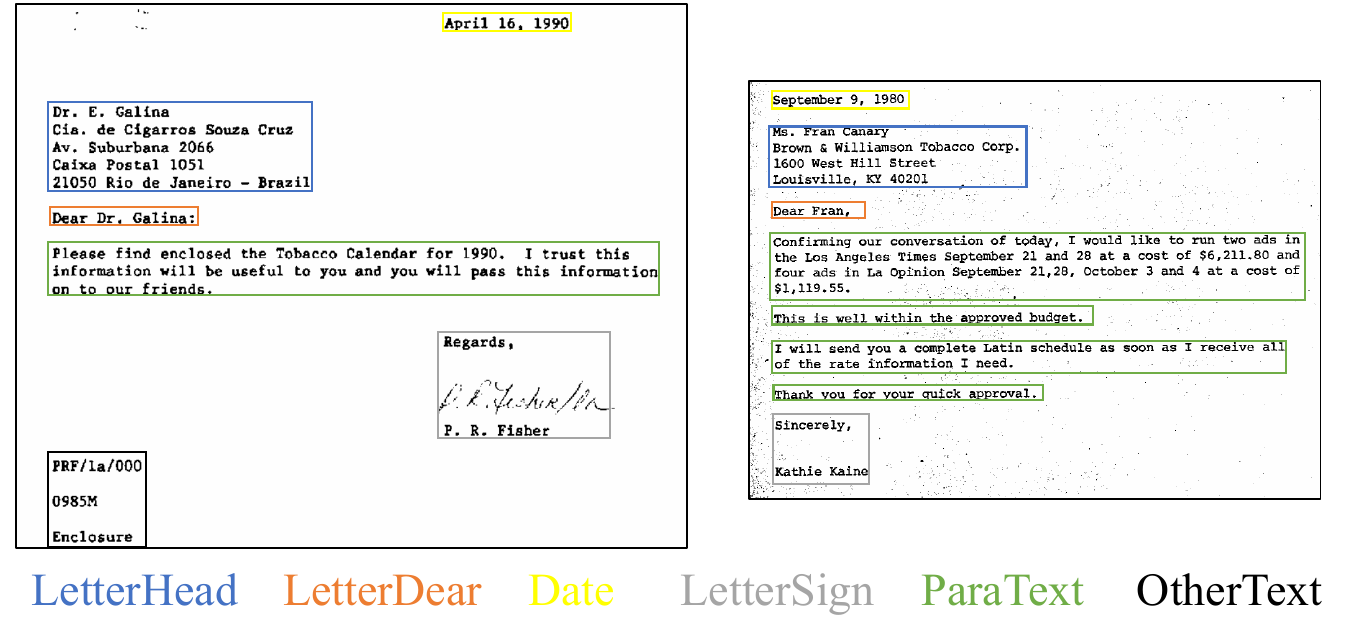}
\caption{Some special layout categories of letters in D$^4$LA. Best viewed in color.}
 \label{fig:vis_letters}
\end{figure}

\noindent
\textbf{Table  and Figure} are common object instances in documents as in PubLayNet and DocBank.

\noindent
\textbf{TableName  and FigureName} are the captions of the Table and Figure, respectively.
While they are both \textit{Caption} in DocBank. 

\noindent
\textbf{Footer} is the footnote of the document, which often begins with special symbols.

\noindent
\textbf{PageHeader and PageFooter} are the page header and page footer on the page, respectively.

\noindent
\textbf{Author} represents the author of the paper or other documents, \eg, News article, Scientific report.

\begin{figure*}[t]\centering
 \includegraphics[width=0.95\textwidth]{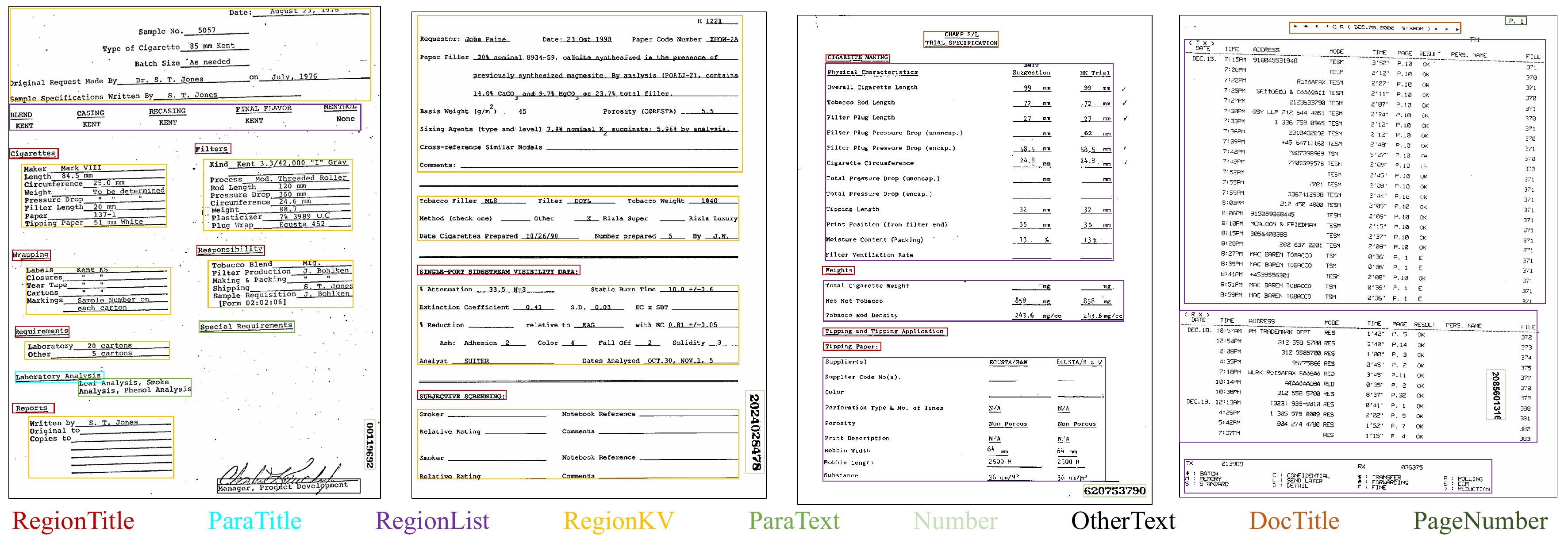}
\caption{Some special layout categories of forms in D$^4$LA. Best viewed in color.}
 \label{fig:vis_kv}
\end{figure*} 

\begin{figure}[!htp]\centering
 \includegraphics[width=0.48\textwidth]{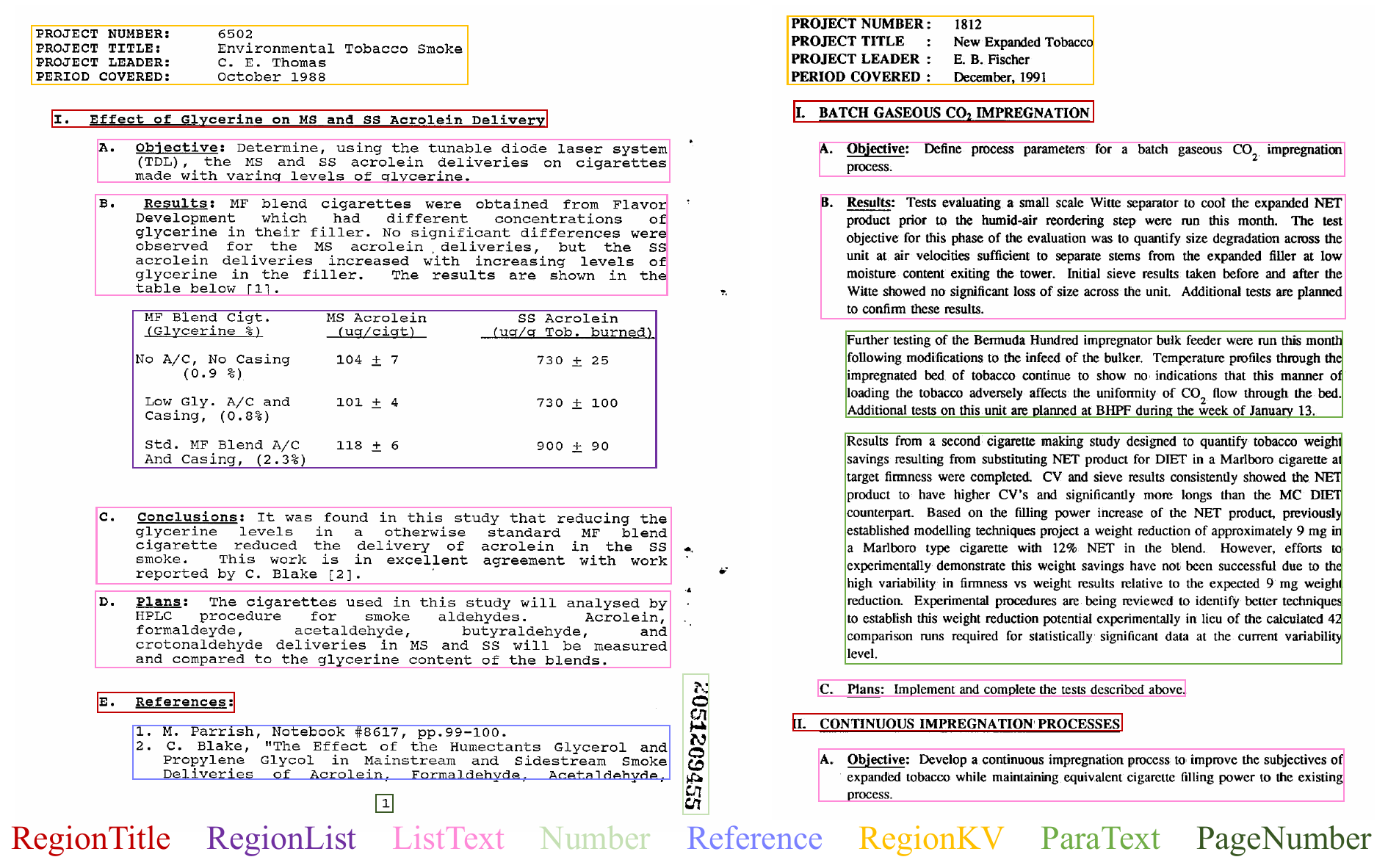}
\caption{Comparison between different types of text.}
 \label{fig:vis_res}
\end{figure} 

\noindent
\textbf{Abstract} often appears at the beginning of the paper behind a section of ``Abstract'' or ``Summary''.

\noindent
\textbf{ParaText} is a paragraph that may have multiple lines when the paragraph is long.
Notably, ``ParaText'' is different from ``ListText'' which contains special enumeration symbols.

\noindent
\textbf{ParaTitle} is similar to \textit{Section} in DocBank, which is the title of one paragraph of ``ParaText''.

\noindent
\textbf{Equation} is the formula or equation in the paper, that often includes formula numbers.

\noindent
\textbf{Reference} often includes a reference number, authors, article name,  journal name, page number, dates, and so on.
All references constitute a ``Reference'' region block.

\noindent
\textbf{PageNumber} is the page number of a document that often appears in the header or footer of the page.

\noindent
\textbf{OtherText} represents some text with word phrases that is not a complete paragraph and is not belong to any other layout categories.
\eg, some useless text.

\subsection{New Categories in Letters} 
By analyzing the documents of letters in RVL-CDIP,  we observe that a standard letter usually has a fixed format.
We customize $3$ classes, \ie, LetterHead, LetterDear and LetterSign for documents of letters, as shown in Figure~\ref{fig:vis_letters}.

\noindent
\textbf{LetterHead} represents the inside address that often appears at the beginning of the letter and records the name and address of the recipient.

\noindent
\textbf{LetterDear} is the salutation or greetings to the recipient, which is usually behind the ``LetterHead''.

\noindent
\textbf{LetterSign} includes the complimentary close and signature, which is often at the end of letters.

\noindent
\textbf{Date} often appears in letters and papers that include years, months, and days.

\begin{figure}[!htp]\centering
 \includegraphics[width=0.48\textwidth]{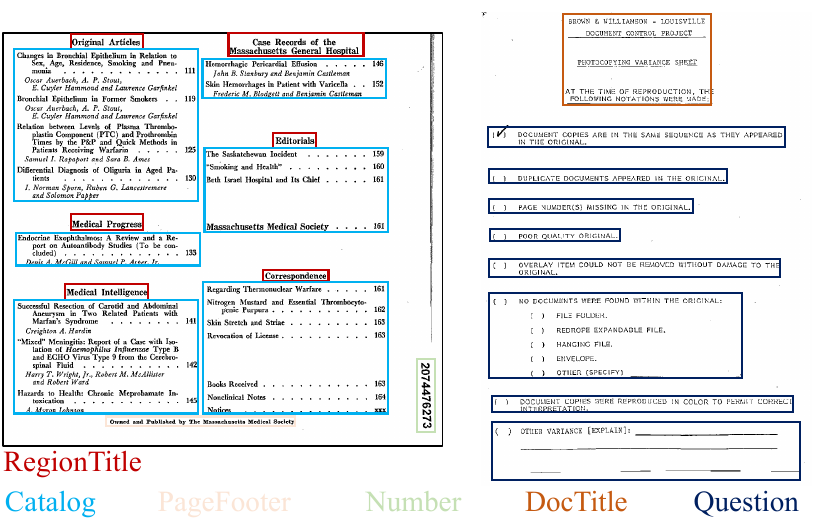}
\caption{Other special layout categories. Best viewed in color.}
 \label{fig:vis_mulu}
\end{figure} 

\subsection{New Categories in Forms} 
Scientific publications are mostly composed of regular paragraphs, tables and figures.
While other documents often include irregular areas, such as the Key-Value pairs in  invoices or the line-less list areas in budget sheets.
This semi-structured data is more important than ordinary words in the document for downstream works, such as information extraction.
Thus, we define $3$ region blocks for special use.
Some cases are illustrated in Figure~\ref{fig:vis_kv}.

\noindent
\textbf{RegionKV} is a region that contains Key-Value areas.

\noindent
\textbf{RegionList} is a region that includes wireless form or line-less list areas.

\noindent
\textbf{RegionTitle} is the title of the complex region, \eg, ``RegionKV'', ``RegionList'' and ``ListText'',
which is different from ``ParaTitle'' of a paragraph.
Typically, both ``ParaTitle'' and ``RegionTitle'' may contain enumeration symbols, which may be confused with ``ListText''.
Thus, distinguishing between these texts requires incorporating the semantics of the context.
We show two difficult cases in Figure~\ref{fig:vis_res}.

\subsection{Other Categories} 
The other remaining categories are shown in Figure~\ref{fig:vis_mulu}.

\noindent
\textbf{Number} represents the special number in IIT-CDIP that is not the content of the document and often vertical text.

\noindent
\textbf{Catalog} includes text and page numbers, which is a region block not one text line with the page number.

\noindent
\textbf{Question} often appears in the questionnaire. They are mostly true or false questions in D$^4$LA dataset.

\end{document}